\documentclass{article}

\usepackage{arxiv}

\usepackage[utf8]{inputenc} 
\usepackage[T1]{fontenc}    
\usepackage{hyperref}       
\usepackage{url}            
\usepackage{booktabs}       
\usepackage{amsfonts}       
\usepackage{nicefrac}       
\usepackage{microtype}      
\usepackage{graphicx}
\usepackage{doi}

\usepackage{pslatex}
\usepackage{subfigure}
\usepackage{float} 
\usepackage{xcolor}         
\usepackage{amsmath}

\title{Online Training of Hopfield Networks using Predictive Coding}

\date{May 19, 2022}	

\author{%
    \href{https://orcid.org/0009-0007-6372-3539}{\includegraphics[scale=0.06]{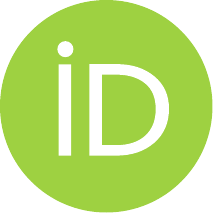}
    \hspace{1mm}Ehsan Ganjidoost} \\
    \texttt{eganjido@uwaterloo.ca}
   \And
   \href{https://orcid.org/0000-0002-9052-2835}{\includegraphics[scale=0.06]{orcid.pdf}
    \hspace{1mm}Mallory Snow} \\
   \texttt{m5snow@uwaterloo.ca}
   \And
   \href{https://orcid.org/0000-0002-4897-8951}{\includegraphics[scale=0.06]{orcid.pdf}
    \hspace{1mm}Jeff Orchard} \\
   \texttt{jorchard@uwaterloo.ca}
   \AND
    Cheriton School of Computer Science\\ Neurocognitive Computing Lab \\
    University of Waterloo,
    Waterloo, ON, Canada \\
}

\newcommand{\pcDynamics}{
\begin{align}
    \tau \frac{d\varepsilon_i}{dt} &= v_i - M_{i}^T \sigma(v_{i+1})  - b_i - \zeta \varepsilon_i \label{errorDynamic} \\
    \tau \frac{d v_i}{dt} &= -\varepsilon_i + W_{i-1}^T \varepsilon_{i-1} \odot \sigma'(v_{i}) \label{valueDynamic}\\
    \gamma \frac{dM_{i}}{dt} &= \varepsilon_{i} \otimes  \sigma(v_{i+1}) \label{predictionDynamic}\\
    \gamma \frac{dW_{i}}{dt} &= \sigma(v_{i+1}) \otimes \varepsilon_{i} \label{correctionDynamic}\\
    \gamma \frac{d b_i}{dt} &= \varepsilon_i \label{biasDynamic}
\end{align}
}

\newcommand{\hopfieldRule}{
\begin{align}
    W &= \frac{1}{N} X^{T}X \ , \label{HopWeights} \\
    b &= \frac{1}{N}\sum_{n=1}^{N} X_n \ , \label{HopBias}  
\end{align}
}

\newcommand{\hopfieldValue}{
\begin{equation}
v_{i} = \begin{cases}
\phantom{-}1 &\text{if } \sum_{j\neq i}W_{ij}v_j\geq 0 \ , \\
-1 &\text{if } \sum_{j\neq i}W_{ij}v_j<0 \ .
\end{cases} \label{HopUpdate}
\end{equation}
}

\newcommand{\hopfieldEnergyOne}{
\begin{equation}
    E = -\frac{1}{2} \sum_{j\neq i}v_iW_{ij}v_j - \sum_jb_jv_j \ . \label{HopEnergy}\\
\end{equation}
}

\newcommand{\hopfieldEnergyTwo}{
\begin{equation}
    E = -\frac{1}{2} \sum_{n}^{N} F(X_n v)- \sum_jb_jv_j \ , \label{HopEnergy2} 
\end{equation}
}

\newcommand{\hyperparametersTable}{
\begin{table}
  \caption{Hyperparameters Setting used for Experiments}
  \label{hyperparameters}
  \centering
  \begin{tabular}{cc|c|c}
    \toprule
    \multicolumn{2}{c|}{Time Constants} & Error Decay & Time Step  \\
    $\tau$ & $\gamma$  & $\zeta$ & $dt$\\
    \midrule
    $v$: 0.1 & $M$: 0.2 & 1 & 1 ms          \\
    $\varepsilon$: 0.1 & $W$: 0.2 & &  \\
                       & $b$: 0.1 &  & \\
    \bottomrule
  \end{tabular}
\end{table}
}

\newcommand{\networkConfigTable}{
\begin{table}
  \caption{Network Configurations used for Experiments}
  \label{netConfig}
  \centering
  \begin{tabular}{ccc}
    \toprule
    \# Layers & \# PC units & Connectivity \\
    \midrule
    3 & (50-30-20) & Loop \\
    1 & (100) & Single Layer \\
    \bottomrule
  \end{tabular}
\end{table}
}

\begin{document}

\maketitle

\begin{abstract}


Neuroscience and Artificial Intelligence (AI) have progressed in tandem, each contributing to our understanding of the brain, and inspiring recent developments in biologically-plausible neural networks (NNs) and learning rules. Predictive coding (PC), and its learning rule, have been shown to approximate error backpropagation in a biologically relevant manner, with local weight updates that depend only on the activity of the pre- and post-synaptic neurons. Unlike traditional feedforward NNs where the flow of information goes in one direction, PC models mimic the brain more accurately by passing information bidirectionally: prediction in one direction, and correction/error in the other. PC models learn by clamping some neurons to target values and running the network to equilibrium. At equilibrium, the network calculates its own error gradients right at the location where they are used for weight updates. Traditional backprop requires the computation graph to be feedforward. However, the PC version of backprop does not have this requirement. Amazingly, no one has demonstrated the application of PC learning directly to recurrent neural networks (RNNs). Hopfield networks (HNs) are RNNs that implement a content-addressable memory, learning patterns (or ``memories'') that can be retrieved from partial or corrupted patterns. 
In this paper, we show that a HN can be trained using the PC learning rules without modification. To our knowledge, this is the first time PC learning has been applied directly to train a RNN, without the need to unroll it in time. 
Our results indicate that the PC-trained HNs behave like classical HNs. 

\end{abstract}
\section{Introduction}
Computational neuroscience aims to understand behavioural or cognitive phenomena at the level of the neuron, or network of neurons. At the same time, artificial neural networks can be trained using the error backpropagation algorithm \cite{rumelhart1986learning}. Recent work has combined these two fields, investigating biologically-inspired alternatives to error backpropagation (backprop).

However, backprop is based on computations that do not have an obvious biological analog. There needs to be a biologically-plausible mechanism that can deliver the relevant output error to the synapses, where biochemical processes can turn that error signal into synaptic weight changes -- learning. This is known as the \emph{credit assignment problem}.



Different approaches have lead to learning methods that respect various biological constraints. It was shown that feedback connections using random weights was sufficient to deliver an error signal that could be used to train the forward weights \cite{lillicrap2016random, nokland2016direct, lillicrap2020backpropagation}. Some approaches use a two-phase learning cycle that extracts a learning signal from the difference between the two resulting network states \cite{scellier2017equilibrium, carreira2005contrastive, xie2003equivalence}. Other methods are able to compute their own error gradients using a specific network architecture  \cite{guerguiev2017towards, Bogacz2017} or neuron model \cite{richards2019dendritic, guerguiev2017towards}.

The biological alternative that we will focus on here is Predictive Coding\footnote{The bidirectional model is backed by different cortical theories \cite{mumford1992computational, rao200216, spratling2010predictive}, and free-energy principals \cite{friston2009predictive}.} (PC), a neural model that is capable of implementing error backpropagation in a biologically-plausible manner \cite{Bogacz2017, Millidge2020, Whittington2019}, on any given topology such as associative memory \cite{salvatori2021associative}.

One of the main factors that makes PC biologically plausible is that the weight update rule requires only local information. That is, the change to the strength of a synapse is based only on the activity of its pre-synaptic and post-synaptic neurons.

PC networks are recurrent, but connections between populations of neurons are still directional. For example, populations $A$ and $B$ can be connected such that $B$ sends a prediction to $A$, while $A$ sends a correction (error) back to $B$.  In this way, layers of neurons can be stacked in a hierarchical network, with predictions flowing in one direction, and corrections flowing in the opposite direction. This directionality allows PC networks to mimic feedforward (FF) neural networks.

Backpropagation can only be used on a FF network because the computation graph is a directed acyclic graph (DAG). So backprop cannot be used directly on PC networks. Instead, PC networks are run to equilibrium, and that equilibrium state distributes the error gradients throughout the network, allowing the update rule to use only local information.

One would think that such a local update rule would be an obvious choice for training RNNs, where backprop does not apply (since the computation graph has cycles). However, no one has yet demonstrated that the PC update rule can be applied directly to RNNs.

In this paper, we apply the PC update rule to a type of RNN, a Hopfield network (HN). We then verify that these PC-trained HNs mimic the behaviour of traditional HNs.

\section{Methods}
Here, we present the mathematical framework of the PC-trained HN. But first, we take a more in-depth look at PC networks and Hopfield networks separately.

\subsection{Predictive Coding}

The concept of predictive coding is built on the idea that the brain's primary function is to minimize prediction error \cite{rao1999predictive}. A Predictive Coding network splits each traditional node (neuron) into two parts, a {\bf value node}, and an {\bf error node}. Together, these two pieces are called a \emph{PC unit}, analogous to a single traditional artificial neuron. These PC units can be connected together and constructed into layers, similar to ANNs. However, in PC networks, predictions (from the value nodes) are sent in one direction through the network, while the prediction errors (held in the error nodes) are sent in the other direction. 

\begin{figure}[b]
\centering
\includegraphics[scale=0.75]{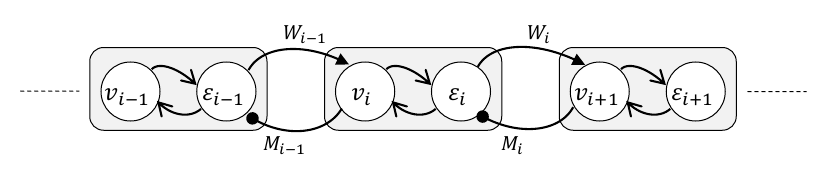}
\caption{Part of a traditional PC network. Each circle represents a population of neurons.}
\label{PC}
\end{figure}
A sample PC network is shown in Fig.~\ref{PC}. Let $v_{i}$ and $\varepsilon_{i}$ denote the value node and error node, respectively, for layer $i$. Then $v_{i}$ sends its prediction down to the layer below, while $\varepsilon_{i}$ receives predictions from the value node of the layer above (i.e., from $v_{i+1}$). In return, $\varepsilon_{i}$ sends up the discrepancy between its value node, $v_i$, and the prediction received from the higher layer. The value node at the higher layer, $v_{i+1}$, takes this discrepancy signal as a correction and adjusts its value accordingly. These downward and upward signals are carried via prediction weights, $M_i$, and correction weights, $W_i$, respectively.

We can write the dynamics of the network using the differential equations,
\pcDynamics
where $\sigma$ is an activation function, $\odot$ is the Hadamard (element-wise) product, $\otimes$ is the outer product, $\zeta$ is a decay coefficient, and $\tau$ and $\gamma$ are time constants in which $\tau < \gamma$. The values used for $\tau$ and $\gamma$ could be different for each of the different above 
equations.  

To train a PC network, one clamps a set of value nodes to their desired values, and runs the rest of the network to equilibrium. Since $\tau<\gamma$, the value nodes and error nodes (equations (\ref{errorDynamic}) and (\ref{valueDynamic})) will reach their equilibrium states before $M$, $W$, and $b$. The states of the value and error nodes are used by equations (\ref{predictionDynamic}), (\ref{correctionDynamic}), and (\ref{biasDynamic}) to update the connection weights and biases. During testing, $M$, $W$, and $b$ do not change, which is equivalent to setting $\gamma$ to infinity.

If the prediction is perfect, the error signal goes to zero, and thus the correction does not make any further changes to the value node. This equilibrium state minimizes the Hopfield-like energy function \cite{Bogacz2017},
\begin{equation} \label{eq:PC_energy}
E = \sum_i \frac{\zeta}{2} \| \varepsilon_i \|^2 \ .
\end{equation}

\subsection{Hopfield Networks}

In an influential paper in 1982, John Hopfield introduced the method of using a recurrent neural network as a content-addressable memory (CAM) by learning patterns (or ``memories'') that can be retreived by intializing the network with a partial or corrupted pattern \cite{hopfield1982neural}.

A Hopfield nework (HN) is a neural network designed to converge to the closest of a finite set of target patterns. In this sense, a HN can be thought of as a dynamical system with a finite number of stable (or equilibrium) states corresponding to the desired target patterns.

In a HN, each neuron, $v_i$, is connected to all other neurons, but not to itself. The connection weights are represented by a symmetric matrix, $W$, where the entry $W_{ij}$ is the weight from $v_{j}$ to $v_i$. Consider a matrix, $X$, which stores a different target pattern in each of its $N$ rows, each row containing $M$ elements. The weights and biases in a HN with $M$ neurons are then given by
\hopfieldRule
where $X_n$ is the $n^{\text{th}}$ row of $X$ (i.e the $n^{\text{th}}$ target pattern).
If we initialize the neurons in the network with some vector of values, each neuron ($v_i$) then changes/updates its value according to the update rule, \hopfieldValue
Hopfield recognized that, like other physical systems, HNs could be described by an energy function, $E$, which takes the form \hopfieldEnergyOne
The first term in the energy function can be interpreted as a measurement of conflict or inconsistency between connected neurons, while the second term represents a ``cost'' for each neuron being active. Some simple rearranging allows us to write the energy in terms of the interaction function, $F$,
\hopfieldEnergyTwo
where $F(x) = x^2$. It is important to point out that the update rule for the neurons in (\ref{HopUpdate}) can be achieved by performing gradient decent on $E$. That is, the algorithm for updating $v$ causes $E$ to monotonically decrease, and the states continue to change until a local minimum is reached.


The interaction function, $F$, traditionally took the form $F(x) = x^2$. For such a HN with $d$ neurons, the number of target patterns it can successfully recall (i.e.\,its storage capacity) is approximately $0.138d$ \cite{amit1985storing}. A lot of work has gone into improving the storage capacity (among other aspects) by using different interaction functions. In more recent years, Modern HNs were introduced using interaction functions of the form $F(x) = x^{n}$ for $n\in \mathbb{Z}_{+}$ \cite{krotov2016dense}, which were shown to have a storage capacity of $d^{n-1}$, as well as improved robustness to adversarial input \cite{krotov2018dense}. An interaction function of the form $F(x) = e^x$ was shown to have a capacity of $e^{\alpha d}$ for some $0 < \alpha < \frac12 \log 2$ \cite{Demircigil2017}.

\subsection{Recurrent Predictive Coding} \label{PC-HNsection}



\begin{figure}[tb]
    \centering
    \includegraphics[width=0.30\textwidth]{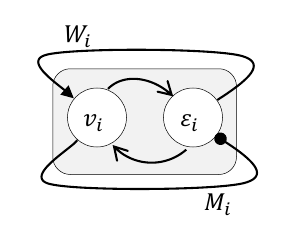}
    \caption{A single population of PC units, recurrently connected to itself. The circles represent populations of neurons. The arrows internal to the unit are one-to-one connections, while the connections that loop outside the unit are dense connections.}
\label{SinglePC}
\end{figure}
\begin{figure}[tb]
    \centering
    \includegraphics[width=0.4\textwidth]{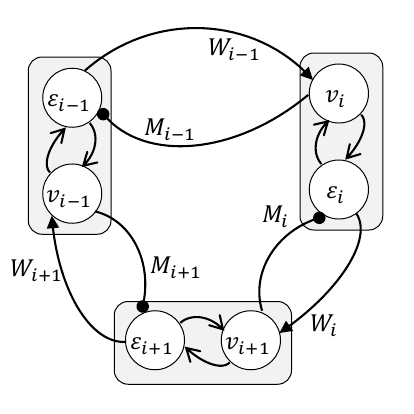}
    \caption{A loop of three PC populations. Each PC population is not recurrently connected to itself, but the entire loop forms a recurrent network.}
\label{LoopPC}
\end{figure}

We can construct a HN out of PC units by connecting the PC layers recurrently. We will call this kind of network a Predictive-coding Hopfield network (PC-HN). We will investigate two different architectures: one that is a single population of 100 PC units connected recurrently, and another that is made up of 3 layers connected in a loop (with 50, 30 and 20 PC units in each respective layer). Figure~\ref{SinglePC} shows the single population PC network, and Fig.~\ref{LoopPC} shows the loop of PC layers. A summary of these network configurations is shown in Table~\ref{netConfig}.
\networkConfigTable

The beauty of this arrangement is that we do not need to implement the HN update rule. Instead, we construct the PC network and have it use its own PC learning rules (equations (\ref{predictionDynamic})-(\ref{biasDynamic})) to update the connection weights and biases.
After training, we initialize the value nodes by setting them to a certain state, then the value and error nodes continually update according to equations (\ref{errorDynamic}), and (\ref{valueDynamic}).

If we train this PC-HN on a set of $N$ patterns, we expect that it will behave as a traditional HN by being able to recall these target patterns from a corrupted version of the target.

\section{Experiments}
We ran a series of experiments on the two different PC-HN architectures described in section \ref{PC-HNsection} (single population of 100 units, and a loop of populations with 50, 30, and 20 units). The set of target patterns consisted of 10 patterns, randomly generated using two different distributions: real-valued, and binary. The real-valued targets were constructed by sampling target patterns from a Gaussian distribution, $\mathcal{N}(0, I)$, while the binary targets were constructed from a binomial distribution, $\{-1,   1\}$, with equal probability. Both architectures have 100 units, so the target patterns were vectors of 100 values. For the networks trained on the real-valued targets, we used the ReLU activation function. For the networks trained on the binary targets, we used the $\tanh$ activation function.

During training, all 100 value nodes were clamped to their target values for 10 seconds. During that time, the other dynamic variables continued to be governed by equations (\ref{predictionDynamic}), (\ref{correctionDynamic}), and (\ref{biasDynamic}). During testing (in the experiments listed below), however, no nodes were clamped. Instead, the experiments investigate the state that the network converges to, and how it depends on the initial state of the network. 

To monitor if the network was converging to one of the target states, we measured the distance from the network state to each of the target states. For the real-valued targets, we used Euclidean distance. For the binary targets, we used the Hamming distance (counting the number of mismatched bits).

The goal of a Hopfield network is to converge to a sensible target pattern when initialized in a nearby state. In other words, when a Hopfield network is initialized with a perturbed or corrupted target pattern, the network should converge back to the corresponding target, essentially correcting the errors in the initial state.

The following experiments aim to verify that our PC-trained HNs also have this quality. We will show that the target patterns are equilibrium states of the PC-HN, and nearby initial states converge to the targets. All experiments were run on a simple desktop computer; training a network on 13 targets took less than a minute. The hyperparameter values we used are show in Table~\ref{hyperparameters}. The code is available upon request.
\hyperparametersTable


\subsection{Linear Stability Analysis}
To show analytically that the target patterns correspond to equilibrium states, we performed a linear stability analysis, forming the Jacobian matrix of the dynamical system linearized at each equilibrium. We verified that the real part of the eigenvalues of the Jacobian matrix are all negative. Interestingly, most of the eigenvalues were $\frac{-1}{2\tau}$; recall that $\tau$ is the time constant from (\ref{errorDynamic}) and (\ref{valueDynamic}). There was always an eigenvalue close to -1, and a small number of eigenvalues that were on the order of $10^{-2}$ to $10^{-4}$.

\subsection{Perturbation Study}
Next, we initialized the network with a perturbed version of a target, and then ran the network to convergence. If the system converged back to the target state, then that initial state was in the basin of attraction of the target, indicating that the network was able to correct the corrupted initial pattern.

For the real-valued targets, we perturbed the targets with noise sampled from a Gaussian distribution, $\mathcal{N}(0, 0.5I)$, while the binary targets were perturbed by ``flipping'' 13 bits of the target (i.e., changing the sign of 13 randomly chosen bits). We initialized the PC-HN with these perturbed targets and measured the distance between the network state and the target as the network ran.

\begin{figure}[tb]
    \centering
    \subfigure[50-30-20 Real-valued]{
        \includegraphics[width=2.6in]{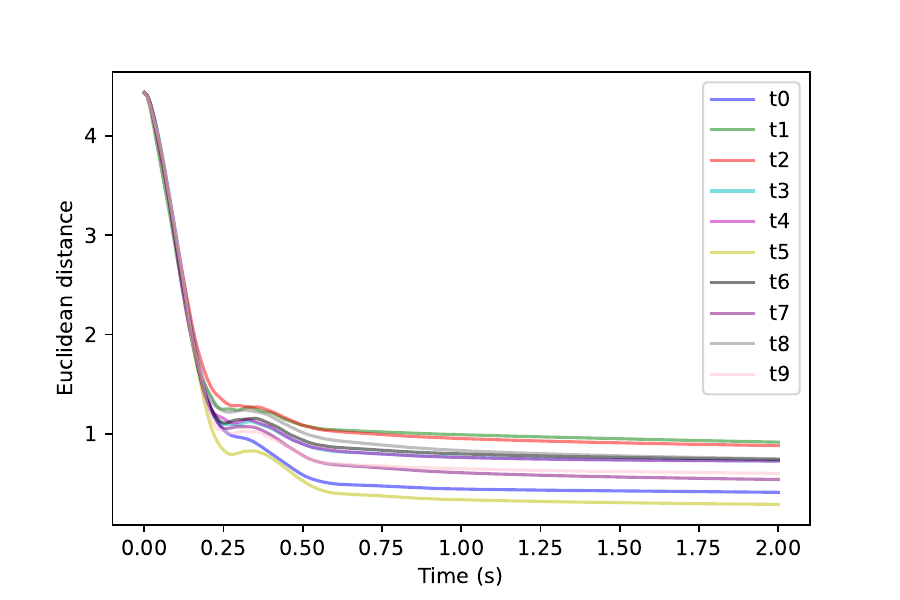}
    }
    \subfigure[50-30-20 Binary]{
        \includegraphics[width=2.6in]{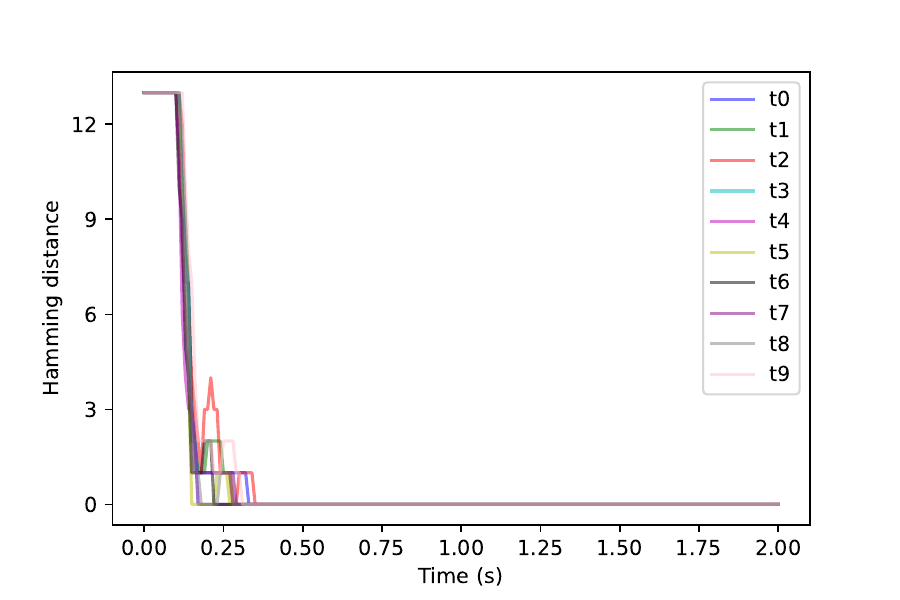}
    }\\
    \subfigure[100 Real-valued]{
        \includegraphics[width=2.6in]{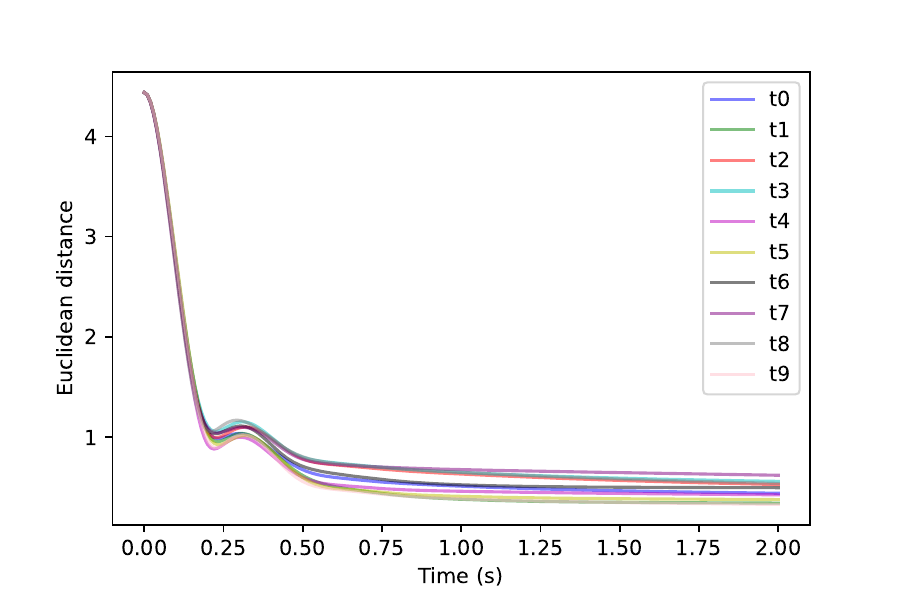}
    }
    \subfigure[100 Binary]{
        \includegraphics[width=2.6in]{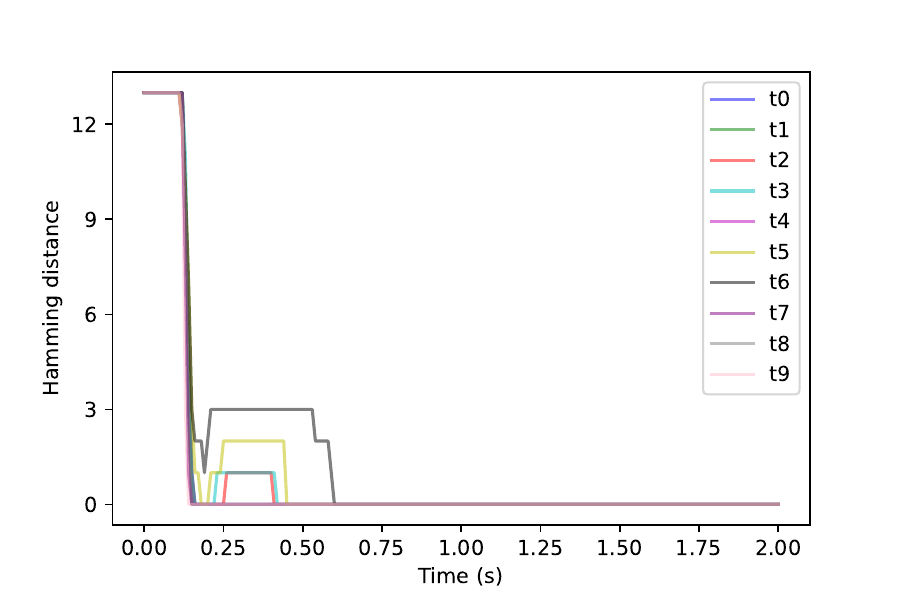}
    }
    \caption{Each plot shows the distance between the network state and the target patterns over time after being initialized with perturbed targets. This was done for each of the 10 targets the network was trained on (see legends) and in all cases, the the network converges to the target pattern. Results are presented from the 50-30-20 loop (top) and the single population of 100 units (bottom), and for real-valued targets (left) and binary targets (right).}
    \label{Exp1}
\end{figure}

Figure \ref{Exp1} presents these distances collected from both network architectures (the 50-30-20 loop, and single population of 100 units), and both target types (real-valued, and binary). All 4 plots show that the distance to the corresponding target decreases as the network evolves through time. This shows that the networks were able to correct the corrupted patterns, and recall the corresponding target state with high fidelity.

\begin{figure}[tb]
    \centering
    \subfigure[Real-valued]{
        \includegraphics[width=.45\textwidth]{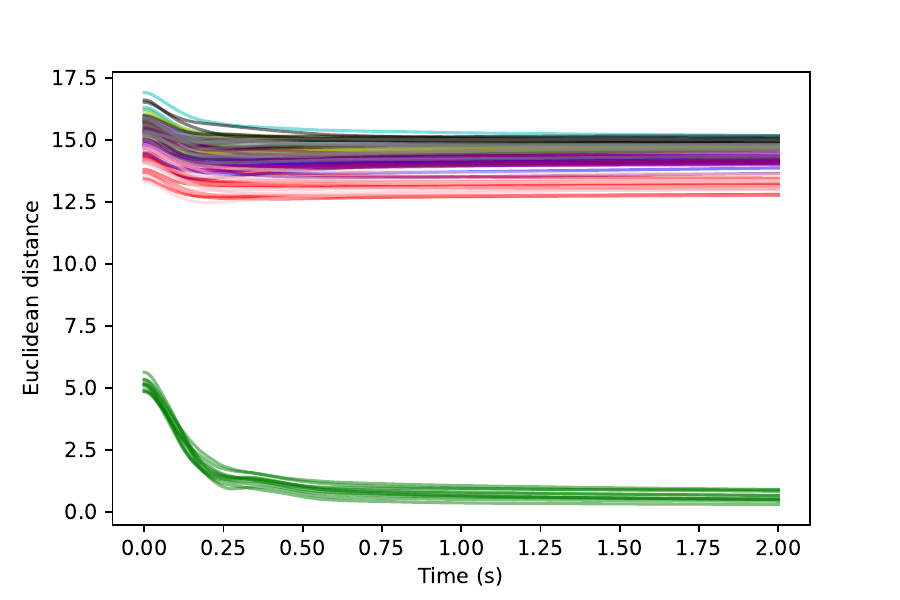}
    }
    \subfigure[Binary]{
        \includegraphics[width=0.45\textwidth]{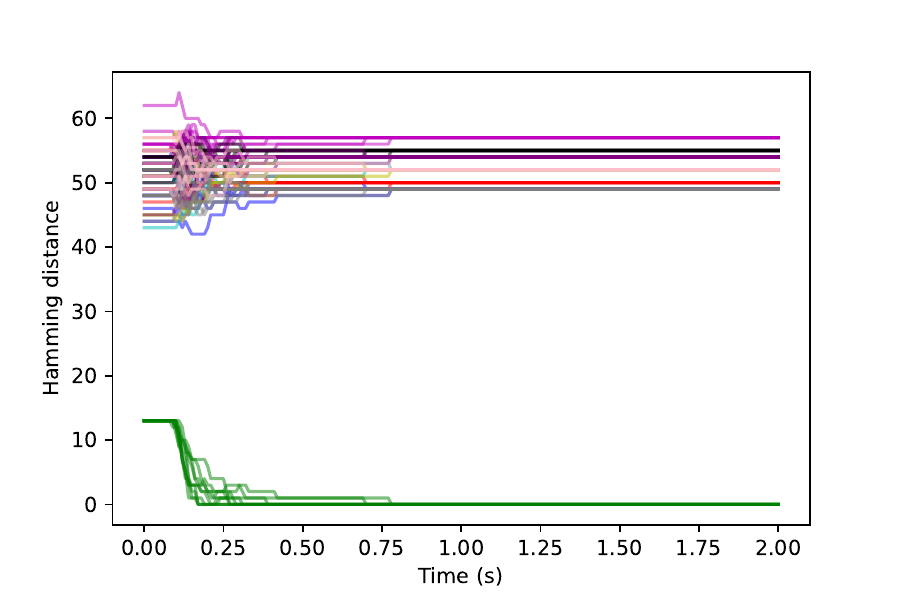}
    }
    \caption{Ten runs showing convergence to the corresponding target, and not to any of the other targets. Each plot includes 10 lines of each colour, where the colour indicates the distance to one of the target patterns. Notice that the network converges to target 1 (green), and does not converge to any of the other targets. These plots show results for the 50-30-20 network, but results were similar for the single-population network, and for other targets.}
    \label{fig:all_in_one}
\end{figure}

Figure \ref{fig:all_in_one} shows that while the network converges close to the corresponding target state, it does {\bf not} converge to any of the other target states. We ran our network initialized from 10 perturbations of target 1, and calculated the distance of the evolving network state from each of the targets. As we would expect, the distance to target 1 decreases. But the figure shows that the distances to the other targets do not decrease, indicating that the network state is converging only to the target state that it was initialized near.

\subsection{Random Initial States}

\begin{figure}[tb]
    \centering
    \subfigure[Real-valued]{
        \includegraphics[width=.45\textwidth]{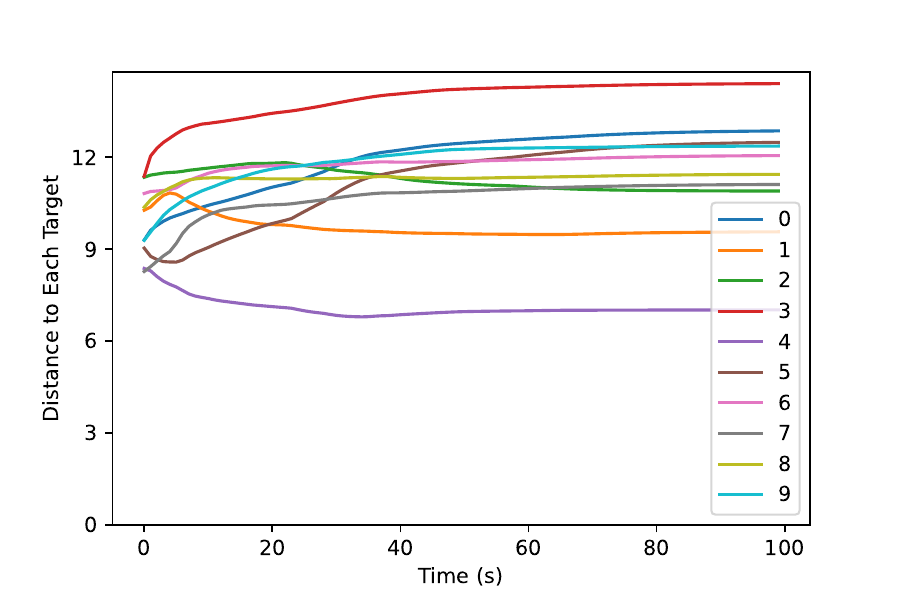}
    }
    \subfigure[Binary]{
        \includegraphics[width=0.45\textwidth]{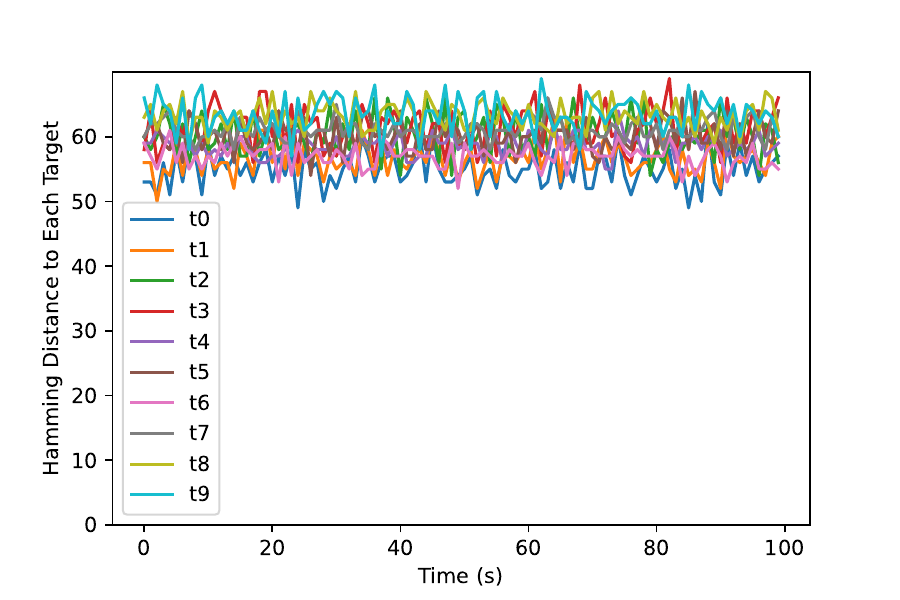}
    }
    \caption{Network behaviour starting from a random initial state. Notice that the network state does not converge to any of the target states. These plots show results for the 50-30-20 network, but results were similar for the single-population network.}
    \label{fig:random_inits}
\end{figure}

Finally, we ran some experiments starting from random initial states. This is different from the experiments above, which were based on perturbations of targets. Starting with a random initial state, fully unhinged from any of the targets, the network usually did not converge to one of the targets. Instead, in the case of the network trained on real-valued targets, the network tended to converge slowly to some other (non-target) equilibrium states. In the case of the network trained on binary targets, the network seemed to oscillate chaotically around some nondescript state. Figure~\ref{fig:random_inits} shows runs starting from initial states, 10 runs for the network trained on real-valued targets, and 10 runs for the network trained on binary targets.

\section{Discussion}
The perturbation study shows that the target patterns correspond to equilibrium states. We initialized the PC-HN with perturbed versions of the targets, and monitored the distance between the state of the network and the correct target patterns. In Fig.~\ref{Exp1}, we can see that --  in all cases -- the distance to the correct target decreases substantially toward zero. This indicates that the network corrects these perturbations as it converges to the equilibrium corresponding to the target pattern. 

The linear stability analysis verified that the target patterns are stable equilibrium states. More specifically, we linearized the dynamical system and evaluated the Jacobian around the equilibrium points, then verified that the real part of Jacobian's eigenvalues are all negative. However, some of the equilibrium states had eigenvalues that were near zero. Convergence along the direction of the corresponding eigenvectors could be slow. This phenomenon might be apparent in our experiments; in Fig.~\ref{Exp1}(a) and (c), convergence seems to slow considerably for some of the runs, indicating that they might be approaching the target from a slower direction.

The system seems to contain some spurious equilibrium states that do not correspond to target patterns. Running the network from a random initial state resulted in very slow convergence to an unknown equilibrium (see Fig.~\ref{fig:random_inits}). As we know, running a HN corresponds to gradient descent on the Hopfield function. We hypothesize that the Hopfield function has a number of shallow, local minima that the network can get trapped in.
\section{Limitations and Future Work}
\label{Limitations_FutureWork}
This work takes a significant step forward in applying PC learning to RNNs, but some aspects remain for future investigation. For example, one could extend this work to Modern HNs. It is not clear how the activation functions and energy function of the PC network could be chosen to implement the modern HN dynamics, but it warrants investigation.

In our linear stability analysis, the real parts of a number of the eigenvalues were close to zero. Why this happens, and what might be done to avoid it, remains a topic of future investigation.

We wonder if training the PC-HN with noisy targets might ameliorate the issue with near-zero eigenvalues. This remains to be seen.

Finally, Hopfield networks are only the ``tip of the iceberg'' for recurrent neural networks. It would be very interesting to apply the PC learning strategies to a wider family of RNNs, such as those used in NLP.


\section{Conclusion}
Both PC and HNs have received considerable attention for their biologically plausible properties. Learning in PC is based on the hypothesis that the brain is continually working to minimize prediction error. This learning mechanism has been shown to be a more biologically sound approximation to back propagation that does not require the computation graph of a given network to be feedforward. Hence, an intuitive step is to apply PC learning to RNNs. Hopfield networks are RNNs that also have interesting biological qualities -- namely, they provide a basis for modelling human memory as they store memories that can be elicited by partial or noisy information. This work presents the first application of PC learning to HNs, and the first use of PC networks to train RNNs without unrolling the network in time. 
\section*{Acknowledgements}
This research was supported by funding from the Natural Sciences and Engineering Research Council (NSERC) of Canada. We also gratefully acknowledge the support of NVIDIA Corporation for the donation of a Titan Xp GPU used for this research.


\end{document}